\documentclass{article}

\usepackage{microtype}
\usepackage{graphicx}
\usepackage{subfigure}
\usepackage{booktabs} 

\usepackage{hyperref}
\usepackage[table,xcdraw]{xcolor}
\definecolor{lightorange}{rgb}{1, 0.8, 0.4}
\definecolor{lightyellow}{rgb}{1, 1, 0.6}
\definecolor{lightpink}{rgb}{1, 0.8, 0.9}

\usepackage[accepted]{icml2025}

\usepackage{amsmath}
\usepackage{amssymb}
\usepackage{mathtools}
\usepackage{amsthm}
\usepackage{enumitem}
\usepackage[capitalize,noabbrev]{cleveref}

\theoremstyle{plain}

\theoremstyle{definition}

\theoremstyle{remark}

\usepackage[textsize=tiny]{todonotes}

\icmltitlerunning{PhiP-G: Physics-Guided Text-to-3D Compositional Scene Generation}

\begin{document}

\twocolumn[
\icmltitle{PhiP-G: Physics-Guided Text-to-3D Compositional Scene Generation}



\icmlsetsymbol{equal}{*}
\icmlsetsymbol{corresponding}{$\dagger$}
\begin{icmlauthorlist}
\icmlauthor{Qixuan Li}{sch,yyy}
\icmlauthor{Chao Wang}{sch,yyy,corresponding}
\icmlauthor{Zongjin He}{sch,yyy}
\icmlauthor{Yan Peng}{sch,yyy}
\end{icmlauthorlist}

\icmlaffiliation{sch}{School of Future Technology, Shanghai University, Shanghai, 200444, China.}

\icmlaffiliation{yyy}{Institute of Artificial Intelligence, Shanghai University, Shanghai, 200444, China}


\icmlcorrespondingauthor{Chao Wang}{cwang@shu.edu.cn}

\icmlkeywords{ Compositional Scene , ICML}

\vskip 0.3in
]



\printAffiliationsAndNotice{}  

\begin{figure*}[!h]
    \begin{center}
    \centerline{\includegraphics[scale=0.5]{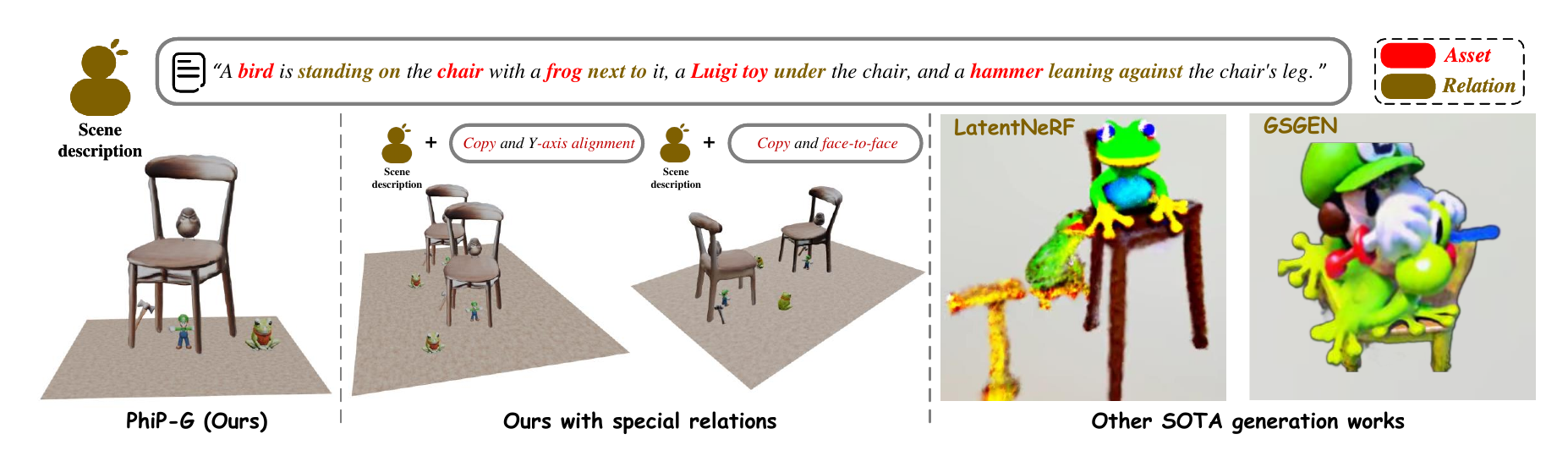}}
    \vskip -0.2in
    \caption{\textbf{PhiP-G} is dedicated to understanding   complex scene descriptions and generating high-quality 3D compositional scenes while supporting the generation of special scene relationships. Compared with existing generation methods, our method demonstrates excellent physical consistency and the ability to handle special environmental relationships.}
    \label{kaiju}  
    \end{center}
    \vskip -0.3in
\end{figure*}

\begin{abstract}
    Text-to-3D asset generation has achieved significant optimization under the supervision of 2D diffusion priors. However, when dealing with compositional scenes, existing methods encounter several challenges: 1).~failure to ensure that composite scene layouts comply with physical laws; 2).~difficulty in accurately capturing the assets and relationships described in complex scene descriptions; 3).~limited autonomous asset generation capabilities among layout approaches leveraging large language models (LLMs). To avoid these compromises, we propose a novel framework for compositional scene generation, \texttt{PhiP-G}, which seamlessly integrates generation techniques with layout guidance based on a world model. Leveraging LLM-based agents, \texttt{PhiP-G} analyzes the complex scene description to generate a scene graph, and integrating a multimodal 2D generation agent and a 3D Gaussian generation method for targeted assets creation. For the stage of layout, \texttt{PhiP-G} employs a physical pool with adhesion capabilities and a visual supervision agent, forming a world model for layout prediction and planning. Extensive experiments demonstrate that \texttt{PhiP-G} significantly enhances the generation quality and physical rationality of the compositional scenes. Notably, \texttt{PhiP-G} attains state-of-the-art (SOTA) performance in CLIP scores, achieves parity with the leading methods in generation quality as measured by the T$^3$Bench, and improves efficiency by 24×.

\end{abstract}
\section{Introduction}
Text-to-3D models \citep{HiFA,dreamfusion,magic3d,SyncDreamer} are systems that convert natural language descriptions into 3D assets by integrating techniques like 2D-to-3D conversion, neural implicit representation, and 3D mesh generation, leveraging deep learning for cross-modal and diverse 3D content creation. There is a growing demand for high-quality 3D assets, particularly for training scenarios in autonomous driving \citep{autod} and robotic navigation \citep{Bermejo2021}. In contrast, 3D content creation, especially for complex scenes, often requires substantial time and effort from domain experts, resulting in constrained production capacity. The advent of text-to-3D technologies offers a novel solution to this challenge, empowering non-expert users to create 3D assets through natural language. However, existing 3D generation methods typically prioritize improving the quality of individual asset, while paying insufficient attention to tasks like compositional scene generation.
 

Compositional scene generation refers to the process of generating a finite number of 3D assets based on scene descriptions and arranging them in a physically plausible layout. Current mainstream text-to-3D models generally lack an understanding of complex semantics and guidance for the layout of scene-level 3D assets. As a result, generating the compositional 3D scene involving multiple objects frequently gives rise to disorganized layouts and inadequate physical consistency. (i.e., \textbf{issue 1}). Concurrently, a recent trend \citep{CompoNeRF,Set-the-Scene,po2023compositional3dscenegeneration} involves manually designed layouts to impose geometric constraints, capturing relationships among multiple objects in the scene, and using implicit neural radiance fields (NeRF) \citep{NeRF} for generation. However, this approach struggles to meet all constraints in the layout, leading to blurry textures and geometric distortions (i.e., \textbf{issue 2}). In contrast, some recent 3D scene generation methods employ LLMs as agents to analyze textual descriptions and leverage the reasoning capabilities of LLMs for layout guidance \citep{SCENECRAFT,Holodeck}. Nevertheless, these models often focus only on 3D asset layouts, requiring assets to be sourced from existing 3D assert libraries. Such limitations inherently constrains their 3D asset generation capabilities, significantly reducing the flexibility of the generation model (i.e., \textbf{issue 3}). From the above issues, 3D compositional scene generation emerges as a task that extends beyond merely stacking assets. This task requires models with exceptional single-asset generation capabilities, advanced semantic understanding, and physics-based layout guidance. The absence of any component can result in catastrophic quality in compositional scene.  As shown in the right panel of Figure \ref{kaiju}, traditional methods lacking physical layout guidance and complex semantic understanding lead to chaotic results.

In this paper, we propose a framework {named \texttt{PhiP-G} (Physics-Guided Text-to-3D Compositional Scene Generation)} for generating high-quality 3D scenes from complex natural language inputs. In \texttt{PhiP-G}, we integrate generative models with LLM-based agents for 3D asset generation. And then we utilize the predictive and planning capabilities of the world model \citep{world1,world2,world3} during the layout phase to construct high-quality 3D scenes that comply with physical laws and align with textual descriptions, without additional training. Specifically, we employ LLM-based agents to perform semantic parsing and relationship extraction on complex textual inputs, generating a scene graph to avoid manual layout by the user (addressing \textbf{issue 2}). 

Simultaneously, we combine a DALL·E 3-based 2D image generation agent with the 3D Gaussian splatting (3DGS) \citep{Splatting} generation model DreamGaussian \citep{DreamGaussian}, incorporating a CLIP-based \cite{clip} score filtering mechanism and a 2D image retrieval library to form the 3D asset generation module. The module enables flexible generation of high-quality assets based on the decomposed scene graph (addressing \textbf{issue 3}). 

We use Blender
as the foundational platform for layout design, introducing a physical pool with a \textit{physical magnet} and a relationship-matching agent for coarse compositional scene layout. A visual supervision agent evaluates coarse layout and provides iterative fine-tuning guidance. These two stages of layout guidance form the world model, demonstrating excellent performance in semantic consistency with complex textual inputs and adherence to physical laws (addressing \textbf{issue 1}). Extensive experiments demonstrate that \texttt{PhiP-G} achieve free, flexible, and physically consistent high-quality 3D compositional scene generation without requiring additional  training.

Our \textbf{contributions} can be summarized as follows:
\begin{itemize}
\item We propose a framework \texttt{PhiP-G}, based on 3DGS for text-to-3D generation and world model-based scene layout, which enables the rapid generation of the high-quality, continuous, and physically consistent 3D compositional scene from textual prompts.
\item \texttt{PhiP-G} enhances the understanding of complex scene descriptions through a multi-agent text preprocessing mechanism, incorporates a physical pool with a physical magnet, and leverages world model attribute prediction to improve the physical coherence of compositional scene layout guidance.
\item In extensive 3D composite scene generation experiments, \texttt{PhiP-G} achieves SOTA on the semantic consistency metric CLIP. Particularly, on the T$^3$Bench metric, it matches SOTA in overall performance while improving generation efficiency by 24 times.
\end{itemize}

\section{Related Work}
\begin{figure*}[!h]
\begin{center}
\centerline{\includegraphics[scale=0.55]{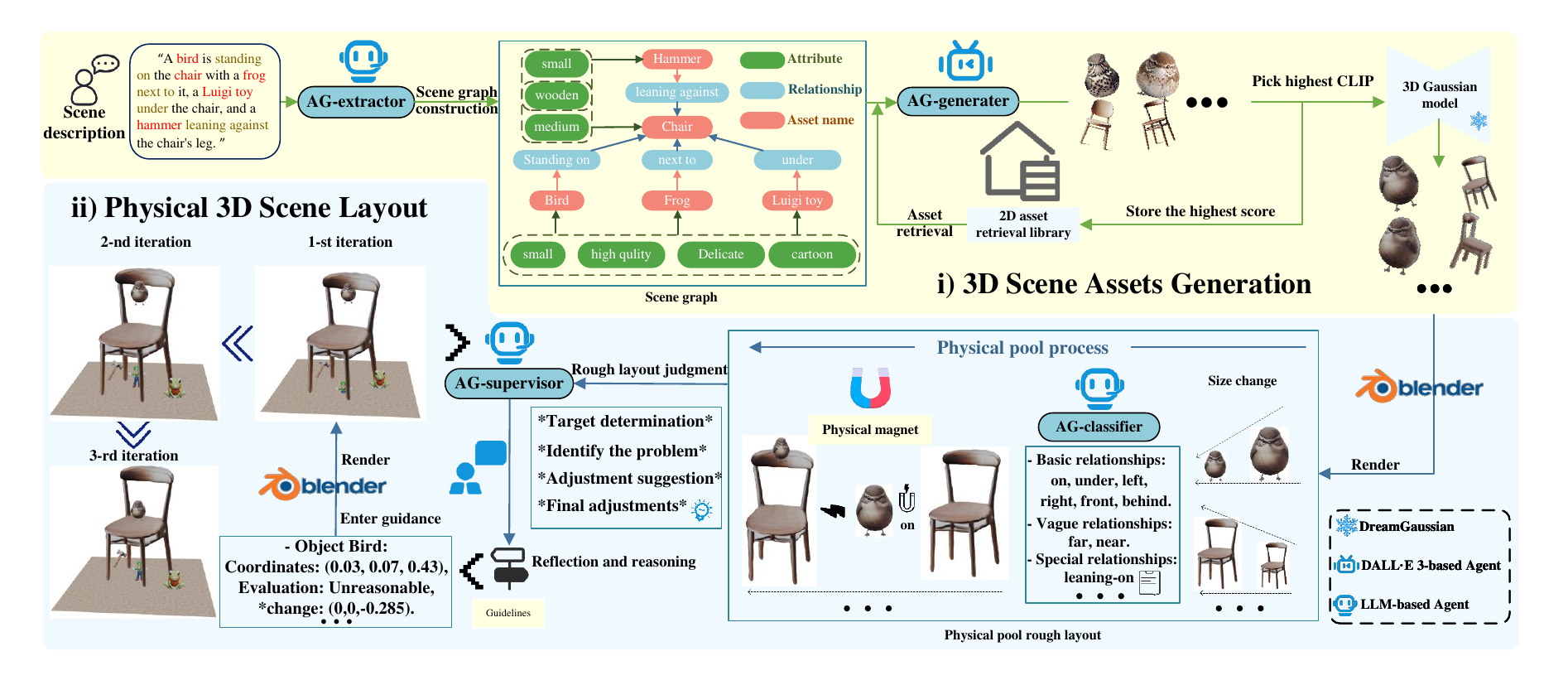}}
\vskip -0.2in
\caption{\textbf{Overview of PhiP-G.} Given a complex scene description, PhiP-G employs an LLM-based agent to perform text analysis and construct a scene graph. Graph-based 3D asset generation is carried out using a 2D generation agent and the 3D Gaussian model, where the 2D asset with the highest CLIP score is stored in the 2D retrieval library for future use. 
Subsequently, Blender serves as the foundational environment, where a  world model consisting of the physical pool and a visual supervision agent enables coarse layout and iterative refinement. PhiP-G ensures improved semantic consistency and physical coherence in the generated scene.}
\label{overview}  
\end{center}
\vskip -0.4in
\end{figure*}
\textbf{3DGS for text-to-scene generation.}
Traditional text-to-3D methods primarily rely on generative approaches based on adversarial networks or variational autoencoders \citep{gan1,gan2,ferreira2022gan,VAE1,VAE2,petrovich2021action}, utilizing 2D images or textual descriptions to infer and generate complex 3D shapes, which are often computationally expensive and slow to produce. The recently popular 3DGS \citep{Splatting} demonstrates a method for representing 3D spaces by optimizing 3D Gaussian spheres, enabling fast rendering and making it popular in 3D scene reconstruction. \citep{chung2023luciddreamer} generates 3D scenes through image inpainting with stable diffusion \citep{stablediffusion}, using reference images or text to expand different viewpoints. GALA3D \citep{Gala3d} utilizes object-level text-to-3D modeling, and MVDream \citep{MVDream} generates realistic objects, combining them using scene-level diffusion models. Text-to-scene generation methods based on 3DGS excel in specific tasks but struggle with consistency and stability in complex, detail-rich scenes. Moreover, their training demands substantial computational resources. Our approach integrates 3DGS with LLM-based scene layout, utilizing 3DGS for individual asset generation to enhance speed and stability while ensuring broad compatibility with most 3DGS methods without requiring additional training.

\textbf{Multimodal LLM agents for scene generation.} With the advancement of multimodal LLMs, models equipped with visual perception capabilities, such as GPT-V \citep{openai2023gpt4v}, have become increasingly sophisticated. Consequently, there is a growing interest in integrating multimodal LLMs into 3D scene generation tasks. For instance, Holodeck \citep{Holodeck} employs a multi-stage process to transform initial 3D scene layouts derived from text into realistic environments. The process uses ChatGPT-4 \citep{openai2023gpt4v} for spatial reasoning, layout generation, material selection, and object arrangement, optimizing spatial relationships to achieve realistic 3D interactions. Similarly, SceneCraft \citep{SCENECRAFT} utilizes a modular architecture with LLMs to iteratively convert textual descriptions into 3D spatial layouts, object selection, and attribute settings, enabling interactive 3D world creation and code generation from natural language instructions. However, most LLM-based scene generation methods rely on LLMs for reasoning, code generation, and asset retrieval, but are limited by their inability to independently create 3D assets and their dependence on pre-existing libraries. In contrast, our approach fully leverages the advanced multimodal capabilities of LLMs and integrates a 3DGS model, allowing for the free generation of target 3D assets from scene descriptions.

\section{Method}

\subsection{Overview}

As illustrated in Figure \ref{overview}, our overall framework consists of two main components: 1).~3D assets generation, and 2).~physical 3D scene layout. For the first component, a keyword extraction agent infers the scene graph and relationships from a complex scene description $T$, followed by the generation of 3D assets using a 2D image generation agent and a 3D Gaussian generation model (Section \ref{3DG}). For the second component, the framework incorporates a designed physical pool with a relationship classification agent and an iterative fine-tuning visual supervision agent as part of the world model, enabling a two-stage compositional scene layout (Section \ref{Layout}).

\subsection{{Generation of High-Quality 3D Scene Assets}}\label{3DG}
This section mainly introduces the process of generating 3D scene objects from scene descriptions. First, we explain how the scene descriptions are processed. Then, we describe the use of the treated text in generating 3D scene assets.

\textbf{Scene description processing.} When our brain receives a description of a complex scene, we instinctively think about the objects in the scene and their sizes and relationships. Analogously, when constructing a 3D scene from a description, a ``brain" is required to process it. Therefore, we design a keyword extraction agent, \texttt{AG-extractor}, which uses the Chain of Thought (CoT) reasoning \citep{wei2023chainofthoughtpromptingelicitsreasoning} method to decompose complex scene descriptions and generate a scene graph $G(\cdot)$, as shown in Figure \ref{overview}. First, for the scene description statement \( T \), we use the agent to decompose and extract all assets in the scene \( A = \{a_i \mid i \in [1, \dots, k]\} \), where \( k \) is the total number of scene assets. The descriptions are then enriched and refined to obtain asset text descriptions more suitable for 2D generation, represented as $ \hat{A} = \{\hat{a}_i \mid i \in [1, \ldots, k]\} $. Next, the keyword extraction agent evaluates the size of each extracted asset $a_i$ and generates an asset size classification $Z = \{ z_i \mid i \in [1, \ldots, k] \}$. We categorize asset sizes into three types: large, medium, and small.

After reasoning the size of each asset, keyword extraction agent performs basic relationship reasoning based on each asset's name $a_i$, size $z_i$, and their relationships within the scene description. To avoid the extensive spatial relationship training required in previous layout methods, we sort the assets according to their extraction order.
The agent selects the second asset as the core asset and infers the relative spatial relationships between each asset and the core asset. These relationships are described as  
$R = \{ r_i \mid i \in [1, \ldots, k] \}$. 
Since the core asset corresponds to $i = 2$, its relationship $r_2$ is set to \texttt{None}.
Such method allows for initial positioning and incremental scene layout based solely on the object order and their relationships to the core asset. As a result, no additional training is needed. After determining the basic relationships, we infer the special relationships for the overall scene, such as duplication requirements and alignment strategies. The special relationships are denoted as $S \in \{ s_1, s_2, s_3 \}$, where, $s_1$, $s_2$, and $s_3$ represent the $x$-axis, $y$-axis, and face-to-face alignment, respectively. 
At this point, we successfully obtain the scene graph $G(T) = (S, A, Z, R)$ for the complex scene description. The entire formal expression of the extractor can be shown as $\texttt{AG-extractor}(T) \rightarrow \big(S, A, Z, R) = G(T)$. where, \( A \), \( Z \), and \( R \) respectively represent all the extracted assets in the scene, their corresponding sizes, and their relationships with the core assets. 

\textbf{3D scene assets generation.}
For the 3D scene asset generation part, we use the enriched asset descriptions \( \hat{A} \) as generation prompts. To ensure both generation quality and speed, we divide the 3D asset generation into two parts: 1).~using a 2D generation agent to convert the asset descriptions into 2D images, and 2).~rapidly generating 3D assets from the 2D images through a Gaussian generation model.

For the first part, we design a 2D generation agent, \texttt{AG-generater}, which is based on DALL·E 3 to generate 2D images \( I_{\text{2D}} = \{i_{\text{2D}}^{j} \mid j \in [1, \dots, k]\} \) from textual descriptions \( \hat{A} \). Due to the excessively rich training samples of DALL·E 3, the 2D generation process often produces abundant backgrounds and elements outside the descriptions, which can adversely affect the subsequent 3D generation. Therefore, we further use prompt engineering on \texttt{AG-generater} to constrain the generation requirements, including specifications for backgrounds, shadows, and material textures of objects. To further ensure semantic consistency between the 2D-generated images and asset descriptions, we adopt a looped generation process to produce multiple images (typically five). 

And then, we calculate their CLIP scores relative to the asset descriptions and select the image \( i_{\text{2DMax}}^{j} \) with the highest text similarity for subsequent 3D generation. The above processing steps ensure the quality and semantic consistency of the generated 2D images, which is also beneficial for subsequent 3D asset generation. The overall formal expression of the 2D generation agent is $ \texttt{AG-generater}(\hat{A}) \rightarrow \big(i_{\text{2DMax}}^{1}),\dots,\big(i_{\text{2DMax}}^{k})$. In addition to directly generating images from text, we have established a 2D image retrieval library. When the required asset is present in the retrieval library, the image is invoked directly, bypassing the text-to-2D process of the \texttt{LLM-generator}. If the asset is not found, the image with the highest CLIP score \( i_{\text{2DMax}}^{j} \), is assigned a name based on the asset and stored in the retrieval library. Additionally, the retrieval library enables users to manually name and store specific 2D images, enhancing stability during multiple generations, improving generation efficiency, and eliminating semantic ambiguity related to the need for specific assets in the scene.

For the second part, considering the need for fast 3D asset generation while ensuring quality, we choose the 3D Gaussian generation model. For each scene asset, the 2D image with the highest CLIP score undergoes background removal and repositioning to ensure a clean background and precise object centering. Processed images serve as input to the Gaussian generation model, enabling rapid 3D assets generation. Once the 3D Gaussian assets are obtained, we convert them into the glb format \( I_{\text{3D}} = \{i_{\text{3D}}^j \mid j \in [1, \dots, k]\} \). Compared to the original 3D Gaussian data, glb format is more convenient for downstream tasks and facilitates our subsequent scene layout process. By extracting and transforming assets from scene descriptions into 3D, the scene layout no longer relies on pre-existing 3D data asset libraries, significantly improving the flexibility of the overall compositional scene generation task.

\subsection{Two-Stage Compositional 3D Scene Layout}\label{Layout}
This section primarily describes the two-stage compositional scene layout process based on the world model. In the first stage, we implement an initial scene layout through the design of simple and effective physical pool. In the second stage, we design a scene layout supervision agent that optimizes the scene layout through a feedback loop with supervised learning. The two-stage process together forms the world model framework, enabling the prediction and planning of scene layouts based on existing information and 3D assets.

\textbf{Preliminary layout design.}
In the first stage of the scene layout, based on the scene graph $G(T)$ and the generated 3D scene assets, we designed a physical pool to perform the preliminary layout. Within the physical pool, we define a relational database with rich relationships, classifying basic physical relationships in the real world. And then, we introduce a classification agent called \texttt{AG-classifier} to match the inter-asset relationships with the relational database. The agent identifies these relationships as the standard relationships between 3D assets denoted as \( \hat{R} = \{ \hat{r_i} \mid i \in [1 \ldots k] \}\), as shown in $\texttt{AG-classifier}(R) \rightarrow (\hat{r_1}),\dots,(\hat{r_k})$. This approach standardizes the types of relationships, clarifying abstract relationship descriptions, which facilitates unified management and invocation of spatial relationships.
Due to the fact that the assets generated by the 3D Gaussian generation model are of nearly the same size, we need to uniformly scale the assets based on their inferred sizes so that their dimensions conform to real-world physical laws. Next, we perform coarse bounding box extraction, which involves obtaining the local bounding boxes of the objects and mapping them to the world coordinate system. We displace the 3D assets according to the standardized relationship $ \hat{R}$, such that the bounding boxes are tangential in the corresponding standardized manner. For example, the relationship ``on" means that the bottom of one asset is tangential to the top of another. The above process completes the rapid displacement and preliminary arrangement of the 3D scene assets.

Simple bounding box intersection alone is insufficient. For instance, in the scene where ``a bicycle leans against a tree", predefined bounding boxes may create gaps between objects, failing to accurately capture physical interaction. To address this, the \textit{physical magnet} is designed to apply vector approximation to the nearest points of two assets, enabling the front asset to ``adhere" to the back asset, similar to the behavior of a magnet. Specifically, it functions as follows: we first utilize Alpha-shape to reconstruct the boundary shape of the 3D assets, obtaining the detailed contour of the asset. To reduce excessively dense and meaningless vertices on the detailed contour, we leverage Blender's merge vertex function and the $Decimate$ modifier to simplify the vertices and mesh, thus lightweighting the detailed contour representation. Then, we select pairs of 3D asset objects that have mutual relationships, iterating through the vertices $\mathcal{V}$ of their detailed contours to calculate the distances and directional vectors of their nearest vertices. By using a contact distance threshold $d_{\text{thresh}}$, we determine whether there is contact between these two assets. If no contact exists, we displace the former asset according to the nearest vertex distance and direction vector. 

The \textit{physical magnet} effectively eliminates the empty spaces caused by simple bounding box tangency that violate physical laws, enabling touch-based contact between two assets without requiring training. Under the combined influence of bounding box tangency constraints in asset relationships and the \textit{physical magnet}, a rapid preliminary scene layout is generated, which partially adheres to physical laws. However, this preliminary layout may be insufficient for more complex scenes. The \textit{physical magnet} can lead to violations of physical laws, such as causing a bird to be attracted to the back of the chair when describing a scenario exemplified by ``a bird standing on a chair". The formal expression of the entire \textit{physical magnet} is given as:
\begin{multline} \label{eq:vertex_update}
    \mathcal{V}_{\text{simplified}} = \text{Decimate}(\mathcal{V}_{\text{orig}}), \quad |\mathcal{V}_{\text{simplified}}| < |\mathcal{V}_{\text{orig}}|, \\
    \quad d(v_1, v_2) = \min_{v_2 \in \mathcal{V}_2} \|v_1 - v_2\|, \quad \vec{d}(v_1, v_2) = v_2 - v_1, \\
      v_1 + \lambda \cdot \vec{d}(v_1, v_2) \rightarrow v_1^{\text{new}} \quad \text{if} \quad d(v_1, v_2) > d_{\text{thresh}}.
    \end{multline}
where, \( \mathcal{V}_{\text{orig}} \) and \( \mathcal{V}_{\text{simplified}} \) represent the original and simplified vertex sets, respectively. \( v_1 \) and \( v_2 \) denote the current object's vertex and the nearest vertex of the target object. \( d_{\text{thresh}} \) defines the vertex contact threshold, \( \lambda \) scales the vertex displacement, and \( v_1^{\text{new}} \) represents the updated vertex position for guiding asset displacement.

\textbf{Feedback loop for layout optimization.}
To further optimize the scene layout, we designed a visual supervision loop for iterative improvement. By utilizing a scene supervisor agent with visual capabilities, the \texttt{AG-supervisor}, the preliminary layout is evaluated from the perspective of physical commonsense, and guidance is provided for layouts that do not conform to commonsense. To enable the supervisor agent to better understand the required reasoning process and reduce ineffective reasoning, we introduce a reflective process, i.e., reverse reasoning. In the reverse reasoning process, the final target result is first provided, which is that all objects should be correctly placed. Then, we provide examples and detailed adjustment suggestions, including reasoning for layout adjustments and calculations of displacement distances and directions for the assets. Finally, the framework validates the adjustments to assess the rationality of the move. Such reflection process enhances the adaptability and decision-making quality of the world model in dynamic layout tasks.
\begin{table*}[!t]
    \caption{\textbf{Quantitative analysis of text-to-3D scene layout relies on the CLIP metric.} The advanced methods involved in the comparison are categorized based on model types. \tikz \draw[fill=orange] (0,0) circle (0.1cm); \textbf{SCENE 1}, \tikz \draw[fill=pink] (0,0) circle (0.1cm); \textbf{SCENE 2}, \tikz \draw[fill=magenta] (0,0) circle (0.1cm); \textbf{SCENE 3}, and \tikz \draw[fill=red] (0,0) circle (0.1cm); \textbf{SCENE 4} represent scenes generated using different numbers of assets, with the 3D asset counts being respectively 3, 4, 6, and 8.}
    \label{tb:clip}
    \vskip 0.15in
    \begin{center}
    \begin{small}
    \begin{sc}
    \begin{tabular}{>{\centering\arraybackslash}p{5.5cm}|>{\centering\arraybackslash}p{2.3cm}|>{\centering\arraybackslash}p{1.6cm}|>{\centering\arraybackslash}p{1.6cm}|>{\centering\arraybackslash}p{1.6cm}|>{\centering\arraybackslash}p{1.6cm}}

    \hline
    {\textbf{Methods}} & {\textbf{Model Type}} & \tikz \draw[fill=orange] (0,0) circle (0.1cm); {\textbf{Scene 1}} & \tikz \draw[fill=pink] (0,0) circle (0.1cm); {\textbf{Scene 2}} & \tikz \draw[fill=magenta] (0,0) circle (0.1cm); {\textbf{Scene 3}} & \tikz \draw[fill=red] (0,0) circle (0.1cm); {\textbf{Scene 4}} \\

    \hline
    {LatentNeRF \citep{metzer2022latent}}        & {NeRF}           & 29.68 & 24.96 & 27.87 & 22.59 \\
    {MVDream \citep{MVDream}}           & {NeRF}           & \cellcolor{lightpink}30.72 & 25.04 & \cellcolor{lightpink}28.59 & 25.82 \\
    \hline
    {SJC \citep{sjc}}               & {Voxel Grid}     & 27.05 & \cellcolor{lightyellow}26.10 & 24.45 & 25.62 \\
    \hline
    {DreamFusion \citep{dreamfusion}}       & {3DGS}           & 27.33 & 24.15 & 27.39 & 21.01 \\
    {Magic3D \citep{magic3d}}           & {3DGS}           & \cellcolor{lightyellow}30.46 & 23.06 & 27.89 & \cellcolor{lightyellow}26.83 \\
    {DreamGaussian \citep{DreamGaussian}}     & {3DGS}           & 25.37 & 18.35 & 25.19 & 23.15 \\
    {GSGEN \citep{gsgen}}             & {3DGS}           & 30.28 &\cellcolor{lightpink} 27.40 & \cellcolor{lightyellow}28.41 & \cellcolor{lightpink}30.30 \\
    \hline
    \textbf{{PhiP-G (Ours) }}     & {3DGS + Agents}  & \cellcolor{lightorange} {33.17} & \cellcolor{lightorange}{36.80} & \cellcolor{lightorange}{33.04} & \cellcolor{lightorange}{34.24} \\
    \hline
    \end{tabular}
    \end{sc}
    \end{small}
    \end{center}
    \vskip -0.1in
\end{table*}
In the \( t \)-th iteration, we place cameras along the \( x \)-axis, \( y \)-axis, and \( z \)-axis in the scene layout, and direct them towards the origin to capture simple scene layout reference images from three perspectives, denoted as \( P_t = \{I_X, I_Y, I_Z\} \). We use the three-perspective reference images \( P_t \), the original scene description \( T \), and the scene graph \( G(T) \) as reasoning inputs. The supervisor agent evaluates the rationality of the asset layout and its relationships based on these inputs. The evaluation uses an exact matching mechanism to project complex natural language information into a two-dimensional discrete binary reward score, labeling assets as either ``positive" or ``negative". For assets labeled as ``negative", the \texttt{AG-supervisor} will further provide layout guidance, i.e., based on the asset attributes and the scene, it will offer optimal scene layout suggestions $L_t$. The formal expression of the supervisor is shown as $\texttt{AG-supervisor}(P_t, T, G(T)) \rightarrow L_t$. After the layout optimization is complete, we introduce a scoring function \( S \) to evaluate the layout quality after each optimization guidance, which is used to comprehensively measure the rationality of the layout and the cost of adjustments. The definition is as follows:
\begin{equation} 
S = 1 - \frac{1}{N} \sum_{i \in I_{\text{3D}}} \left( \alpha \cdot \text{Violation}_i + \beta \cdot \frac{|\Delta C_i|}{\Delta_{\text{max}}} \right)
\end{equation}
Here, \( N \) is the total number of 3D assets, and \( \text{Violation}_i \) represents the physical violation degree of asset \( i \). \( \Delta C_i \) denotes the adjustment displacement of asset \( i \), while \( \Delta_{\text{max}} \), the maximum allowable adjustment distance, is defined as \( \Delta_{\text{max}} = 0.5 \). The coefficients \( \alpha \) and \( \beta \) balance the impact of violation degree and adjustment displacement on scoring.

When the layout score \( S_t \) of the \( t \)-th iteration is lower than the previous score \( S_{t-1} \), the current iteration's layout guidance is discarded, and rationality planning is redone. Conversely, if \( S_t \) exceeds the rationality threshold, the layout iteration concludes as reasonable. Otherwise, the layout optimization process repeats until the rationality check is satisfied. Thus, the looped supervision optimization part for the preliminary layout concludes. In the two-stage layout of complex 3D scenes, we fully utilize the predictive planning capabilities of the world model, composed of the physical pool and agents, so that the layout itself does not require training. Compared to traditional layout tasks, this approach saves significant time and computational costs, while yielding satisfactory results.

To enhance the realism of the scene generation, after the layout optimization is completed, we use Blender's built-in particle system to simulate realistic ground surfaces. Three types of realistic ground are simulated: grass, wood, and sand. Meanwhile, we invoke the agent \texttt{AG-extractor} to analyze the scene and determine which type of ground is most suitable. Then, the agent will iterate through the bottom faces of all objects' bounding boxes, identifying the lowest face, and generate the simulated ground at the position where it is tangent to this face.
    \begin{table*}[!h]
        \caption{\textbf{Quantitative analysis of text-to-3D scene layout relies on the T$^3$Bench metric. }The \textbf{Quality} metric evaluates quality and view inconsistency. The \textbf{Alignment} metric measures the consistency between text and 3D scene. The \textbf{Average} is calculated as the average of the quality metric and the alignment metric, reflecting the overall performance of the model.
        }
        \label{tb:t3b}
        \vskip 0.15in
        \begin{center}
        \begin{small}
        \begin{sc}
        \begin{tabular}{>{\centering\arraybackslash}p{6cm}|>{\centering\arraybackslash}p{2.9cm}|>{\centering\arraybackslash}p{1.9cm}|>{\centering\arraybackslash}p{2cm}|>{\centering\arraybackslash}p{1.9cm}}
        \hline
        {\textbf{Methods}} & {\textbf{Running Time} $\downarrow$} & {\textbf{Quality} $\uparrow$} & {\textbf{Alignment} $\uparrow$} & {\textbf{Average} $\uparrow$} \\
        \hline
        {Dreamfusion \citep{dreamfusion}}      & {30mins}    & 17.3 & 14.8 & 16.1 \\
        {Magic3D \citep{magic3d}}          & {40mins}    & 26.6 & \cellcolor{lightyellow}24.8 & 25.7 \\
        {LatentNeRF \citep{metzer2022latent}}       & {65mins}    & 21.7 & 19.5 & 20.6 \\
        {Fantasia3D \citep{chen2023fantasia3d}}       & {45mins}    & 22.7 & 14.3 & 18.5 \\
        {SJC \citep{sjc}}              & {25mins}    & 17.7 & 5.8  & 11.7 \\
        {ProlificDreamer \citep{wang2023prolificdreamer}}  & {240mins}   & \cellcolor{lightorange}45.7 & 25.8 & \cellcolor{lightpink}35.8 \\
        {MVDream \citep{MVDream}}          & {30mins}    & \cellcolor{lightyellow}39.0 & \cellcolor{lightpink}28.5 & \cellcolor{lightyellow}33.8 \\
        {DreamGaussian \citep{DreamGaussian}}    & {7mins}     & 12.3 & 9.5  & 10.9 \\
        {GeoDream \citep{GeoDream1}}         & {400mins}   & 34.3 & 16.5 & 25.4 \\
        {RichDreamer \citep{qiu2024richdreamer}}      & {70mins}    & 34.8 & 22.0 & 28.4 \\
        \hline
       {{PhiP-G (Ours)} }   & {{10mins}} & \cellcolor{lightpink}{42.3} & \cellcolor{lightorange}{29.8} & \cellcolor{lightorange}{36.4} \\
        \hline
        \end{tabular}
        \end{sc}
        \end{small}
        \end{center}
        \vskip -0.2in
    \end{table*}
    \begin{figure*}[!h]
        \begin{center}
        \centerline{\includegraphics[scale=0.55]{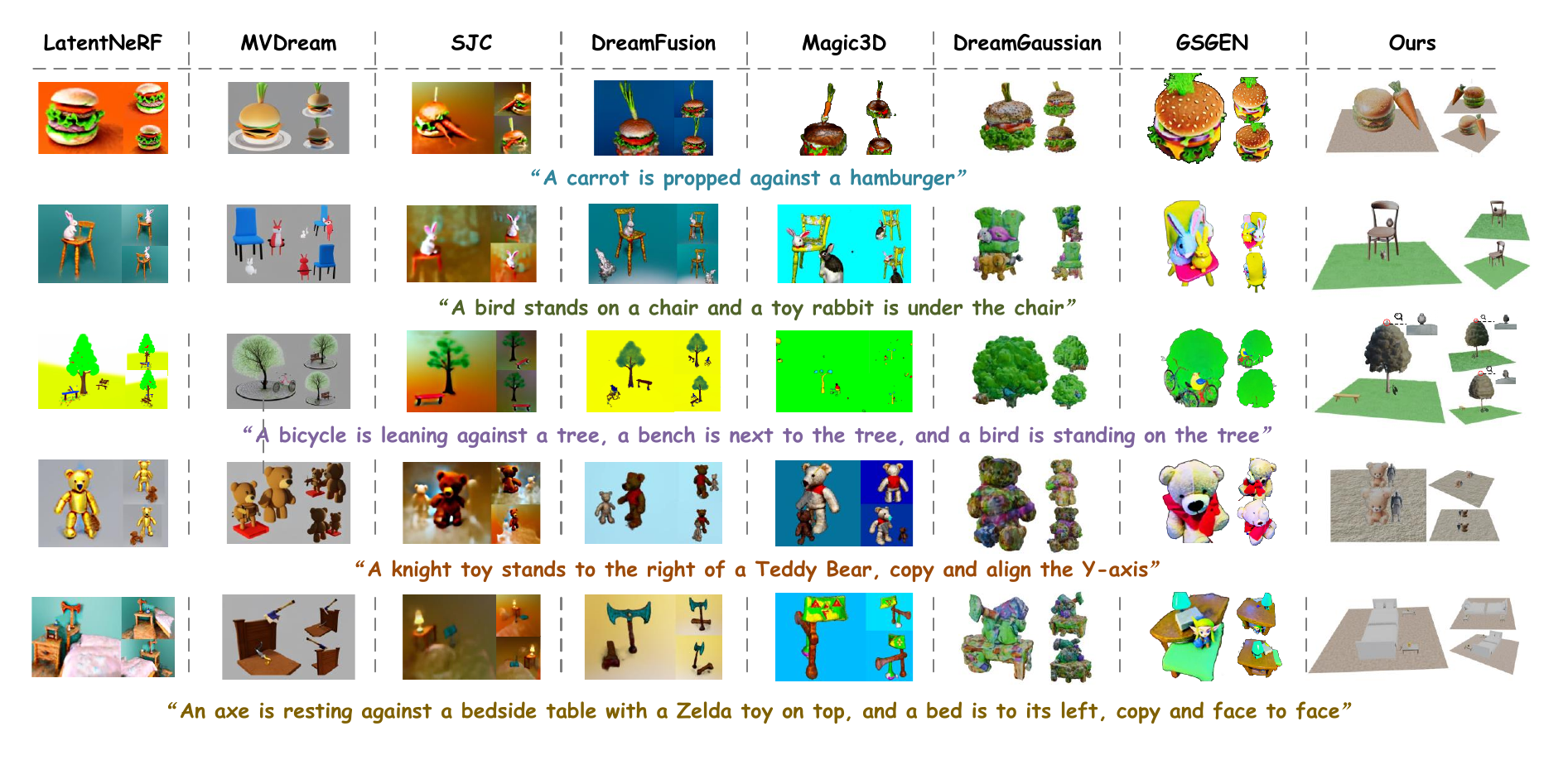}}
        \vskip -0.2in
        \caption{\textbf{Qualitative analysis of text-to-3D scene.} Our method ensures consistency between textual descriptions and generated 3D scenes, while maintaining physical laws and handling special layout requirements.}
        \label{Qualitative_Comparison}  
        \end{center}
        \vskip -0.3in
        \end{figure*}
\section{Experiments}
\subsection{Quantitative Comparison}
\textbf{Quantitative analysis uses the CLIP metric.} In Table \ref{tb:clip}, we use the text-image similarity metric CLIP for qualitative evaluation, analyzing the consistency and quality between text descriptions and 3D scenes. We also compare the performance of our method with current SOTA methods on the task of text-to-3D complex scene construction. To improve the reliability of the benchmark, we consider various 3D reconstruction and representation techniques when selecting advanced methods, including NeRF-driven approaches, voxel-based representation methods, and techniques based on 3D geometric structures. Our approach integrates agents based on the world model while utilizing 3D GS for generation. For a comprehensive demonstration of each model's generation capabilities across diverse 3D scenes, we test scene prompts containing 3, 4, 6, and 8 assets. Ultimately, ours achieves higher CLIP scores than other generation models in the text-to-3D  compositional scene generation and layout task, demonstrating better semantic consistency between text and scenes.

\textbf{Quantitative analysis uses the T$^3$Bench metric.} To further evaluate the generation quality and semantic consistency of ours, we used the evaluation metrics provided by T$^3$Bench \citep{he2023t3bench}, as shown in Table \ref{tb:t3b}. The results of other methods in the table are derived from T$^3$Bench, a comprehensive text-to-3D benchmark specifically designed for evaluating the qulaity of 3D generation. Its quality metrics combine multi-view text-image scoring and regional convolution to detect quality and view inconsistency, while alignment metrics leverage multi-view caption generation and GPT-4 evaluation to measure the consistency between text and 3D content. The evaluation reveals our method achieves the best average performance, with slightly lower quality compared to ProlificDreamer, but a generation time reduced to just one twenty-fourth. The above results fully demonstrates that \texttt{PhiP-G} ensures high scene generation quality while maintaining fast generation speed. Additionally, by utilizing the world model composed of agents, semantic alignment is effectively guaranteed.

\subsection{Qualitative Comparison}
\begin{table*}[!t]
    \caption{\textbf{User study on the results of scene generation.} We conduct a user study to compare our method with the latest related works. Four metrics are selected to evaluate scene quality: text fidelity, scene quality, aesthetics, and physical rationality (PR), where higher scores (on a scale of 0-5) indicate better performance.}
    \label{tb:user_sty}
    \vskip 0.15in
    \begin{center}
    \begin{small}
    \begin{sc}
    \begin{tabular}{>{\centering\arraybackslash}p{5.3cm}|>{\centering\arraybackslash}p{2.3cm}|>{\centering\arraybackslash}p{2.4cm}|>{\centering\arraybackslash}p{2cm}|>{\centering\arraybackslash}p{2cm}}
    \hline
    {\textbf{Methods}} & {\textbf{Text Fidelity}} & {\textbf{Scene Quality}} & {\textbf{Aesthetic}} & {\textbf{PR}} \\
    \hline
    {MVDream \citep{MVDream}} & \cellcolor{lightyellow}{3.41} & \cellcolor{lightyellow}{3.12} & \cellcolor{lightpink}{2.84} & \cellcolor{lightpink}{3.23} \\
    {SJC \citep{sjc}} & \cellcolor{lightpink}{3.36} & 2.99 & \cellcolor{lightyellow}{2.52} & 2.64 \\
    {DreamGaussian \citep{DreamGaussian}} & 2.15 & 2.61 & 2.09 & 2.73 \\
    {GSGEN \citep{gsgen}} & 2.71 & \cellcolor{lightpink}{3.27} & 2.12 & \cellcolor{lightyellow}{3.12} \\
    \hline
    \textbf{{PhiP-G (Ours)}} & \cellcolor{lightorange}{4.55} & \cellcolor{lightorange}{4.32} & \cellcolor{lightorange}{4.87} & \cellcolor{lightorange}{4.95} \\
    \hline
    \end{tabular}
    \end{sc}
    \end{small}
    \end{center}
    \vskip -0.2in
\end{table*}
In the qualitative comparison section, we evaluate our method against SOTA generation models for 3D compositional scene generation performance. To comprehensively demonstrate the quality of the generated scenes, visual displays from the front, back, and side views for each model. As shown in the Figure \ref{Qualitative_Comparison}, our approach demonstrates superior scene generation quality and semantic consistency.

For scenes with more complex textual descriptions, other methods show varying degrees of semantic understanding deviations and missing described assets. Furthermore, when textual descriptions include physical and spatial relationships between assets, others often produce disorganized and misaligned results. In contrast, our approach accurately captures the inter-asset relationships described in the text while maintaining layout rationality and physical consistency in the scene. For specific scene generation requirements, such as ``Copy and face to face", our approach demonstrates clear comprehension and effectively implements these requirements in a logical manner.
\begin{table}[!h]
    \vskip -0.2in
    \caption{\textbf{Scene layout key step ablation.} The term ``w/o LLM-s" refers to ablating \texttt{AG-supervisor}. ``w/o LLM-s \& PM" refers to ablating both \texttt{AG-supervisor} and the \textit{physical magnet}. ``w/o physical pool" refers to ablating the physical pool, while ``w/o All" refers to ablating all the aforementioned components.
    }
    \label{tb:alb}
    \vskip 0.15in
    \begin{center}
    \begin{small}
    \begin{sc}
    \begin{tabular}{>{\centering\arraybackslash}p{3cm}|>{\centering\arraybackslash}p{1.5cm}|>{\centering\arraybackslash}p{1.5cm}}
    \hline
    {\textbf{Methods}} & {\textbf{Time} $\downarrow$} & {\textbf{CILP} $\uparrow$} \\
    \hline
    \textbf{{PhiP-G (Ours)}} & {10mins} & \cellcolor{lightorange}{35.63} \\
    {w/o LLM-s} & {9mins} & \cellcolor{lightpink}{31.93} \\
    {w/o LLM-s \& PM} & {8mins} & 29.55 \\
    {w/o physical pool} & {25mins} & \cellcolor{lightyellow}{30.14} \\
    {w/o All} & {7mins} & 23.25 \\
    \hline
    \end{tabular}
    \end{sc}
    \end{small}
    \end{center}
    \vskip -0.2in
\end{table} 
\begin{figure}[!h]
    \begin{center}
    \centerline{\includegraphics[scale=0.6]{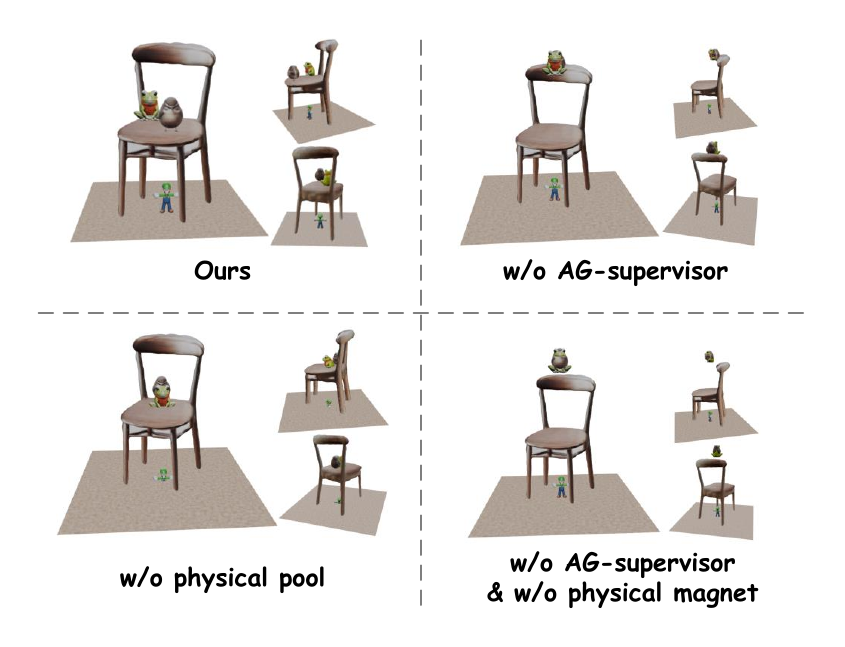}}
    \vskip -0.2in
    \caption{\textbf{Visualization depicting the ablation of key steps.} This ablation experiment visually demonstrates the effectiveness and necessity of each layout module we design.
    }\label{ablation}
    \end{center}
    \vskip -0.3in
\end{figure}
\subsection{Ablation Study}
In Table \ref{tb:alb}, we conduct ablation experiments on the key steps of scene layout. Specifically, we perform ablations in four aspects: 
1).~The intelligent agent \texttt{AG-supervisor} as the layout supervisor. 
2).~Both \texttt{AG-supervisor} and the \textit{physical magnet}. 
3).~The entire physical pool process. 
4).~Removing all three components above simultaneously. The comparison results indicate that our overall layout process effectively balances scene generation time and quality. The absence of \texttt{AG-supervisor} and the \textit{physical magnet} leads to a decrease in generation quality. Moreover, removing the physical pool-based rapid coarse layout significantly increases the optimization iterations required by \texttt{AG-supervisor}, resulting in a substantial increase in fine-tuning time.
Figure \ref{ablation} in presents a visual ablation experiment on key steps. The results show that removing the physical pool forces the supervising agent to handle fine-grained layouts directly, resulting in misaligned assets. When the \texttt{AG-supervisor} is removed, the lack of agent supervision causes coarse layouts through the physical pool without fine-tuning, leading to assets being pulled toward the chair back due to the \textit{physical magnet}.
\subsection{User Study}
To comprehensively evaluate the scene generation quality of our method, we design a user study, as shown in the table \ref{tb:user_sty}. In this experiment, we introduce four evaluation metrics: text fidelity, scene quality, aesthetics, and physical rationality. We invite 73 volunteers to evaluate the scenes generated by our method and other popular models based on these metrics, assigning scores from 0 to 5. Higher scores indicate greater approval of the corresponding aspect of the generated results. From the table, it can be observed that our method performs the best across all metrics, demonstrating that it is more favored by users in scene generation and holds greater potential for further applications. The detailed setup of the overall aforementioned experiments can be found in Appendix \ref{detail}.

\section{Conclusion}
In this paper, we present \texttt{PhiP-G}, a novel text-to-3D compositional scene generation framework that combines advanced 3D Gaussian generation techniques with world model-based layout guidance. The framework excels in generating 3D scenes with strong textual consistency and physical coherence. Extensive experiments validate that \texttt{PhiP-G} outperforms existing methods in compositional scene generation, demonstrating superior semantic understanding and multi-object layout capabilities. Our future work will focus on:  
1).~Incorporating higher-quality 3D generation models into the framework as the 3D generation module;  
2).~Enhancing the integration of world models for more advanced complex scene generation.

\bibliography{example_paper}

\begin{thebibliography}{38}
\providecommand{\natexlab}[1]{#1}
\providecommand{\url}[1]{\texttt{#1}}
\expandafter\ifx\csname urlstyle\endcsname\relax
  \providecommand{\doi}[1]{doi: #1}\else
  \providecommand{\doi}{doi: \begingroup \urlstyle{rm}\Url}\fi

\bibitem[Bai et~al.(2024)Bai, Lyu, Jiang, Li, Lu, Lin, and Wang]{CompoNeRF}
Bai, H., Lyu, Y., Jiang, L., Li, S., Lu, H., Lin, X., and Wang, L.
\newblock Componerf: Text-guided multi-object compositional nerf with editable 3d scene layout, 2024.
\newblock URL \url{https://arxiv.org/abs/2303.13843}.

\bibitem[Bermejo et~al.(2021)Bermejo, Lee, Chojecki, Przewozny, and Hui]{Bermejo2021}
Bermejo, C., Lee, L.-H., Chojecki, P., Przewozny, D., and Hui, P.
\newblock Exploring button designs for mid-air interaction in virtual reality: A hexa-metric evaluation of key representations and multi-modal cues.
\newblock \emph{Proceedings of the ACM on Human-Computer Interaction}, 5\penalty0 (EICS):\penalty0 194:1--194:26, 2021.
\newblock \doi{10.1145/3457141}.

\bibitem[Chen et~al.(2023)Chen, Chen, Jiao, and Jia]{chen2023fantasia3d}
Chen, R., Chen, Y., Jiao, N., and Jia, K.
\newblock Fantasia3d: Disentangling geometry and appearance for high-quality text-to-3d content creation.
\newblock In \emph{Proceedings of the IEEE/CVF International Conference on Computer Vision (ICCV)}, pp.\  22246--22256, October 2023.

\bibitem[Chen et~al.(2024)Chen, Wang, Wang, and Liu]{gsgen}
Chen, Z., Wang, F., Wang, Y., and Liu, H.
\newblock Text-to-3d using gaussian splatting, 2024.
\newblock URL \url{https://arxiv.org/abs/2309.16585}.

\bibitem[Chung et~al.(2023)Chung, Lee, Nam, Lee, and Lee]{chung2023luciddreamer}
Chung, J., Lee, S., Nam, H., Lee, J., and Lee, K.~M.
\newblock Luciddreamer: Domain-free generation of 3d gaussian splatting scenes.
\newblock \emph{arXiv preprint arXiv:2311.13384}, 2023.

\bibitem[Cohen-Bar et~al.(2023)Cohen-Bar, Richardson, Metzer, Giryes, and Cohen-Or]{Set-the-Scene}
Cohen-Bar, D., Richardson, E., Metzer, G., Giryes, R., and Cohen-Or, D.
\newblock Set-the-scene: Global-local training for generating controllable nerf scenes, 2023.
\newblock URL \url{https://arxiv.org/abs/2303.13450}.

\bibitem[Eguchi et~al.(2022)Eguchi, Choe, and Huang]{VAE2}
Eguchi, R.~R., Choe, C.~A., and Huang, P.-S.
\newblock Ig-vae: Generative modeling of protein structure by direct 3d coordinate generation.
\newblock \emph{PLoS computational biology}, 18\penalty0 (6):\penalty0 e1010271, 2022.

\bibitem[Ferreira et~al.(2022)Ferreira, Li, Pomykala, Kleesiek, Alves, and Egger]{ferreira2022gan}
Ferreira, A., Li, J., Pomykala, K.~L., Kleesiek, J., Alves, V., and Egger, J.
\newblock Gan-based generation of realistic 3d data: A systematic review and taxonomy.
\newblock \emph{arXiv preprint arXiv:2207.01390}, 2022.

\bibitem[Ha \& Schmidhuber(2018)Ha and Schmidhuber]{world1}
Ha, D. and Schmidhuber, J.
\newblock Recurrent world models facilitate policy evolution.
\newblock In \emph{Advances in Neural Information Processing Systems 31}, pp.\  2451--2463. Curran Associates, Inc., 2018.
\newblock \url{https://worldmodels.github.io}.

\bibitem[He et~al.(2023)He, Bai, Lin, Zhao, Hu, Sheng, Yi, Li, and Liu]{he2023t3bench}
He, Y., Bai, Y., Lin, M., Zhao, W., Hu, Y., Sheng, J., Yi, R., Li, J., and Liu, Y.-J.
\newblock T$^3$bench: Benchmarking current progress in text-to-3d generation, 2023.

\bibitem[Kerbl et~al.(2023)Kerbl, Kopanas, Leimkühler, and Drettakis]{Splatting}
Kerbl, B., Kopanas, G., Leimkühler, T., and Drettakis, G.
\newblock 3d gaussian splatting for real-time radiance field rendering.
\newblock \emph{ACM Transactions on Graphics}, 42\penalty0 (4):\penalty0 139:1--139:14, 2023.
\newblock \doi{10.1145/3588432.3591417}.

\bibitem[Ko et~al.(2023)Ko, Cho, Choi, Ryoo, and Kim]{gan2}
Ko, J., Cho, K., Choi, D., Ryoo, K., and Kim, S.
\newblock 3d gan inversion with pose optimization.
\newblock In \emph{Proceedings of the IEEE/CVF Winter Conference on Applications of Computer Vision}, pp.\  2967--2976, 2023.

\bibitem[Kosiorek et~al.(2021)Kosiorek, Strathmann, Zoran, Moreno, Schneider, Mokr{\'a}, and Rezende]{VAE1}
Kosiorek, A.~R., Strathmann, H., Zoran, D., Moreno, P., Schneider, R., Mokr{\'a}, S., and Rezende, D.~J.
\newblock Nerf-vae: A geometry aware 3d scene generative model.
\newblock In \emph{International Conference on Machine Learning}, pp.\  5742--5752. PMLR, 2021.

\bibitem[Kumaran et~al.(2023)Kumaran, Rowe, Mott, and Lester]{SCENECRAFT}
Kumaran, V., Rowe, J., Mott, B., and Lester, J.
\newblock Scenecraft: automating interactive narrative scene generation in digital games with large language models.
\newblock In \emph{Proceedings of the Nineteenth AAAI Conference on Artificial Intelligence and Interactive Digital Entertainment}, AIIDE '23. AAAI Press, 2023.
\newblock ISBN 1-57735-883-X.
\newblock \doi{10.1609/aiide.v19i1.27504}.
\newblock URL \url{https://doi.org/10.1609/aiide.v19i1.27504}.

\bibitem[Lin et~al.(2023)Lin, Gao, Tang, Takikawa, Zeng, Huang, Kreis, Fidler, Liu, and Lin]{magic3d}
Lin, C.-H., Gao, J., Tang, L., Takikawa, T., Zeng, X., Huang, X., Kreis, K., Fidler, S., Liu, M.-Y., and Lin, T.-Y.
\newblock Magic3d: High-resolution text-to-3d content creation.
\newblock In \emph{Proceedings of the IEEE/CVF Conference on Computer Vision and Pattern Recognition (CVPR)}, pp.\  300--309, June 2023.

\bibitem[Liu et~al.(2024)Liu, Lin, Zeng, Long, Liu, Komura, and Wang]{SyncDreamer}
Liu, Y., Lin, C., Zeng, Z., Long, X., Liu, L., Komura, T., and Wang, W.
\newblock Syncdreamer: Generating multiview-consistent images from a single-view image.
\newblock In \emph{International Conference on Learning Representations (ICLR)}, 2024.
\newblock Spotlight.

\bibitem[Ma et~al.(2023)Ma, Deng, Zhou, Liu, Huang, and Wang]{GeoDream1}
Ma, B., Deng, H., Zhou, J., Liu, Y.-S., Huang, T., and Wang, X.
\newblock Geodream: Disentangling 2d and geometric priors for high-fidelity and consistent 3d generation.
\newblock 2023.

\bibitem[Metzer et~al.(2022)Metzer, Richardson, Patashnik, Giryes, and Cohen-Or]{metzer2022latent}
Metzer, G., Richardson, E., Patashnik, O., Giryes, R., and Cohen-Or, D.
\newblock Latent-nerf for shape-guided generation of 3d shapes and textures.
\newblock \emph{arXiv preprint arXiv:2211.07600}, 2022.

\bibitem[Micheli et~al.(2023)Micheli, Alonso, and Fleuret]{world2}
Micheli, V., Alonso, E., and Fleuret, F.
\newblock Transformers are sample-efficient world models.
\newblock In \emph{The Eleventh International Conference on Learning Representations}, 2023.
\newblock URL \url{https://openreview.net/forum?id=vhFu1Acb0xb}.

\bibitem[Mildenhall et~al.(2020)Mildenhall, Srinivasan, Tancik, Barron, Ramamoorthi, and Ng]{NeRF}
Mildenhall, B., Srinivasan, P.~P., Tancik, M., Barron, J., Ramamoorthi, R., and Ng, R.
\newblock Nerf: Representing scenes as neural radiance fields for view synthesis.
\newblock In \emph{European Conference on Computer Vision (ECCV)}, 2020.
\newblock URL \url{https://arxiv.org/abs/2003.08934}.

\bibitem[Mittal(2020)]{autod}
Mittal, V.
\newblock Attngrounder: Talking to cars with attention, 2020.
\newblock URL \url{https://arxiv.org/abs/2009.05684}.

\bibitem[OpenAI(2023)]{openai2023gpt4v}
OpenAI.
\newblock Gpt-4v(ision) system card, 2023.
\newblock URL \url{https://openai.com/research/gpt-4}.
\newblock Version 1.0.

\bibitem[Petrovich et~al.(2021)Petrovich, Black, and Varol]{petrovich2021action}
Petrovich, M., Black, M.~J., and Varol, G.
\newblock Action-conditioned 3d human motion synthesis with transformer vae.
\newblock In \emph{Proceedings of the IEEE/CVF International Conference on Computer Vision}, pp.\  10985--10995, 2021.

\bibitem[Po \& Wetzstein(2023)Po and Wetzstein]{po2023compositional3dscenegeneration}
Po, R. and Wetzstein, G.
\newblock Compositional 3d scene generation using locally conditioned diffusion, 2023.
\newblock URL \url{https://arxiv.org/abs/2303.12218}.

\bibitem[Poole et~al.(2023)Poole, Jain, Barron, and Mildenhall]{dreamfusion}
Poole, B., Jain, A., Barron, J.~T., and Mildenhall, B.
\newblock Dreamfusion: Text-to-3d using 2d diffusion.
\newblock In \emph{International Conference on Learning Representations}, 2023.
\newblock URL \url{https://openreview.net/forum?id=FjNys5c7VyY}.

\bibitem[Qiu et~al.(2024)Qiu, Chen, Gu, Zuo, Xu, Wu, Yuan, Dong, Bo, and Han]{qiu2024richdreamer}
Qiu, L., Chen, G., Gu, X., Zuo, Q., Xu, M., Wu, Y., Yuan, W., Dong, Z., Bo, L., and Han, X.
\newblock Richdreamer: A generalizable normal-depth diffusion model for detail richness in text-to-3d.
\newblock In \emph{Proceedings of the IEEE/CVF Conference on Computer Vision and Pattern Recognition}, pp.\  9914--9925, 2024.

\bibitem[Radford et~al.(2021)Radford, Kim, Hallacy, Ramesh, Goh, Agarwal, Sastry, Askell, Mishkin, Clark, Krueger, and Sutskever]{clip}
Radford, A., Kim, J.~W., Hallacy, C., Ramesh, A., Goh, G., Agarwal, S., Sastry, G., Askell, A., Mishkin, P., Clark, J., Krueger, G., and Sutskever, I.
\newblock Learning transferable visual models from natural language supervision.
\newblock \emph{Proceedings of the 38th International Conference on Machine Learning}, 2021.
\newblock URL \url{https://arxiv.org/abs/2103.00020}.

\bibitem[Rombach et~al.(2021)Rombach, Blattmann, Lorenz, Esser, and Ommer]{stablediffusion}
Rombach, R., Blattmann, A., Lorenz, D., Esser, P., and Ommer, B.
\newblock High-resolution image synthesis with latent diffusion models, 2021.

\bibitem[Shi et~al.(2023)Shi, Wang, Ye, Mai, Li, and Yang]{MVDream}
Shi, Y., Wang, P., Ye, J., Mai, L., Li, K., and Yang, X.
\newblock Mvdream: Multi-view diffusion for 3d generation.
\newblock \emph{arXiv:2308.16512}, 2023.

\bibitem[Tang et~al.(2023)Tang, Ren, Zhou, Liu, and Zeng]{DreamGaussian}
Tang, J., Ren, J., Zhou, H., Liu, Z., and Zeng, G.
\newblock Dreamgaussian: Generative gaussian splatting for efficient 3d content creation.
\newblock \emph{arXiv preprint arXiv:2309.16653}, 2023.

\bibitem[Wang et~al.(2022)Wang, Du, Li, Yeh, and Shakhnarovich]{sjc}
Wang, H., Du, X., Li, J., Yeh, R.~A., and Shakhnarovich, G.
\newblock Score jacobian chaining: Lifting pretrained 2d diffusion models for 3d generation.
\newblock \emph{arXiv preprint arXiv:2212.00774}, 2022.

\bibitem[Wang et~al.(2023)Wang, Lu, Wang, Bao, Li, Su, and Zhu]{wang2023prolificdreamer}
Wang, Z., Lu, C., Wang, Y., Bao, F., Li, C., Su, H., and Zhu, J.
\newblock Prolificdreamer: High-fidelity and diverse text-to-3d generation with variational score distillation.
\newblock \emph{arXiv preprint arXiv:2305.16213}, 2023.

\bibitem[Wei et~al.(2023)Wei, Wang, Schuurmans, Bosma, Ichter, Xia, Chi, Le, and Zhou]{wei2023chainofthoughtpromptingelicitsreasoning}
Wei, J., Wang, X., Schuurmans, D., Bosma, M., Ichter, B., Xia, F., Chi, E., Le, Q., and Zhou, D.
\newblock Chain-of-thought prompting elicits reasoning in large language models, 2023.
\newblock URL \url{https://arxiv.org/abs/2201.11903}.

\bibitem[Yang et~al.(2024)Yang, Sun, Weihs, VanderBilt, Herrasti, Han, Wu, Haber, Krishna, Liu, Callison-Burch, Yatskar, Kembhavi, and Clark]{Holodeck}
Yang, Y., Sun, F.-Y., Weihs, L., VanderBilt, E., Herrasti, A., Han, W., Wu, J., Haber, N., Krishna, R., Liu, L., Callison-Burch, C., Yatskar, M., Kembhavi, A., and Clark, C.
\newblock Holodeck: Language guided generation of 3d embodied ai environments, 2024.
\newblock URL \url{https://arxiv.org/abs/2312.09067}.

\bibitem[Zhang et~al.(2021)Zhang, Yang, and Stadie]{world3}
Zhang, L., Yang, G., and Stadie, B.~C.
\newblock World model as a graph: Learning latent landmarks for planning, 2021.

\bibitem[Zhao et~al.(2022)Zhao, Ma, G{\"u}era, Ren, Schwing, and Colburn]{gan1}
Zhao, X., Ma, F., G{\"u}era, D., Ren, Z., Schwing, A.~G., and Colburn, A.
\newblock Generative multiplane images: Making a 2d gan 3d-aware.
\newblock In \emph{European conference on computer vision}, pp.\  18--35. Springer, 2022.

\bibitem[Zhou et~al.(2024)Zhou, Ran, Xiong, He, Lin, Wang, Sun, and Yang]{Gala3d}
Zhou, X., Ran, X., Xiong, Y., He, J., Lin, Z., Wang, Y., Sun, D., and Yang, M.-H.
\newblock Gala3d: Towards text-to-3d complex scene generation via layout-guided generative gaussian splatting.
\newblock \emph{arXiv preprint arXiv:2402.07207}, 2024.

\bibitem[Zhu et~al.(2024)Zhu, Zhuang, and Koyejo]{HiFA}
Zhu, J., Zhuang, P., and Koyejo, S.
\newblock Hifa: High-fidelity text-to-3d generation with advanced diffusion guidance, 2024.
\newblock URL \url{https://arxiv.org/abs/2305.18766}.

\end{thebibliography}
\bibliographystyle{icml2025}

\newpage
\appendix
\onecolumn
\section{Appendix}
\subsection{Implementation details}\label{detail}
We select DreamGaussian as the 3D Gaussian generation model, where the guidance scale is set to 100. The learning rates for opacity, position, and color are set to \( 5 \times 10^{-2} \), \( 1.6 \times 10^{-4} \), and \( 5 \times 10^{-3} \), respectively. For texture and mesh extraction of 3D assets, we set the geometry learning rate to \( 1 \times 10^{-4} \) and the texture learning rate to \( 2 \times 10^{-1} \), which are used for geometric adjustments of the mesh and enhancing texture details. Regarding the agent design, we use GPT-4 as the core for the natural language analysis and reasoning agents \texttt{AG-extractor} and \texttt{AG-classifier}. For the text-to-2D generation agent \texttt{AG-generater}, DALL·E 3 is used as the core, while for the visual supervision agent \texttt{AG-supervisor}, we select GPT-4o \citep{openai2023gpt4v} for its strong visual understanding capabilities. For capturing scene images used in visual supervision, cameras are placed along each axis, facing the origin, with the distance set to 8. Since our work primarily relies on the designed physical pool and various powerful agents for scene layout, training is not required. High-quality compositional scene generation can be completed in approximately 10 minutes on a 12G NVIDIA 4080 Laptop.

\subsection{Physical Relationship Database  within the Physical Pool}
\begin{itemize}
\item Basic relationships: on, under, left, right, front, behind.  
\item Vague relationships: far, near.  
\item Alignment relationship: center-aligned.  
\item Leaning relationship: leaning-on.  
\item Rotation relationships: facing,rotation.  
\item Special relationships: duplicate\_x\_alignment, duplicate\_y\_alignment,duplicate\_facing.
\end{itemize}
We categorize object relationships into basic, vague, alignment, leaning, and rotational relationships, structuring asset positions in scene descriptions accordingly. To further aid the LLM in understanding overall scene adjustment requirements, we define three special relationships: duplicate\_x\_alignment refers to ``copy and align the entire scene along the x-axis", duplicate\_y\_alignment refers to ``copy and align the entire scene along the y-axis", and duplicate\_facing refers to ``copy and face each other".
\subsection{Primary Agent Prompt}\label{a_prompt}
\begin{figure*}[!t]
    \begin{center}
    \centerline{\includegraphics[scale=0.75]{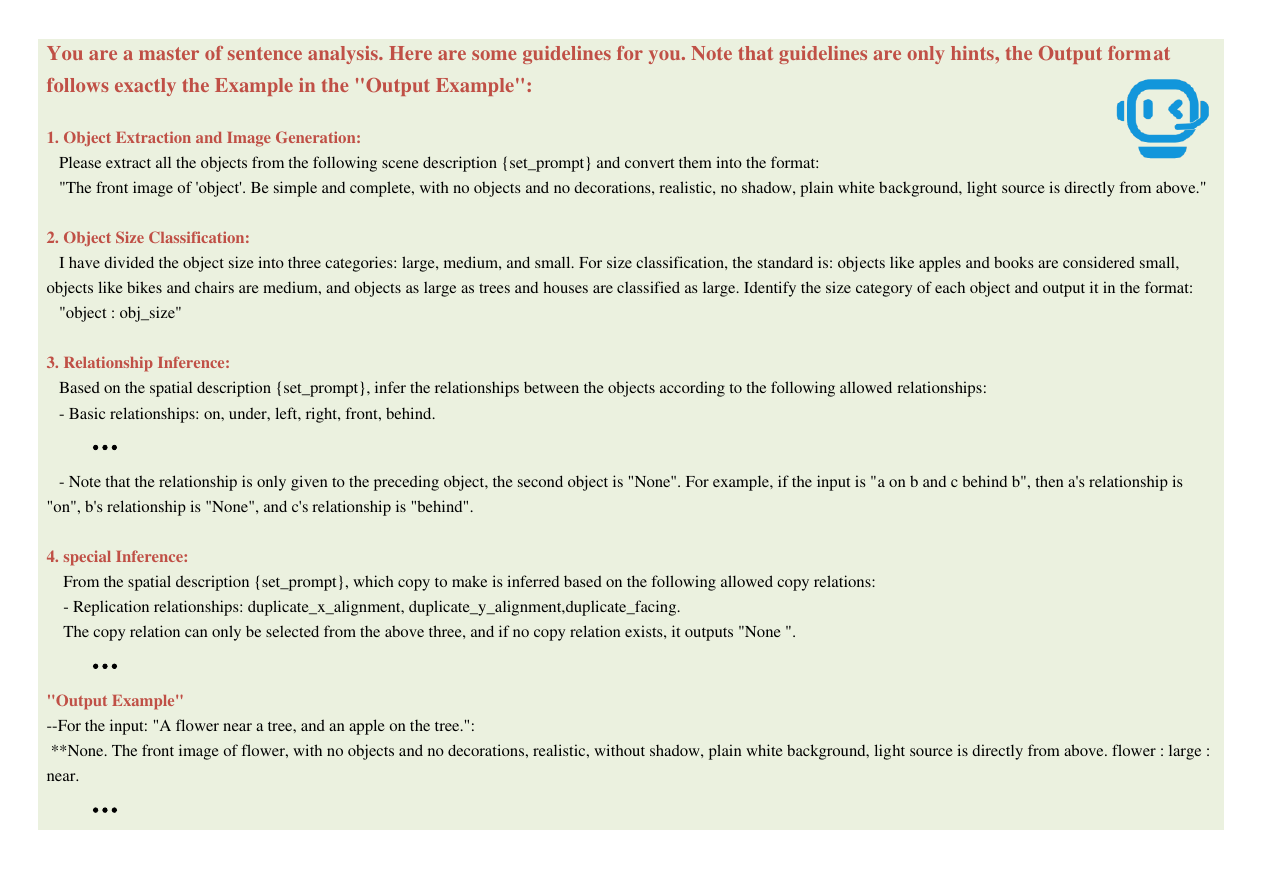}}
    \vskip -0.2in
    \caption{Example of agent AG-extractor prompt.}
    \label{prompt}  
    \end{center}
    \vskip -0.3in
\end{figure*}

\begin{figure*}[!t]
        \begin{center}
        \centerline{\includegraphics[scale=0.75]{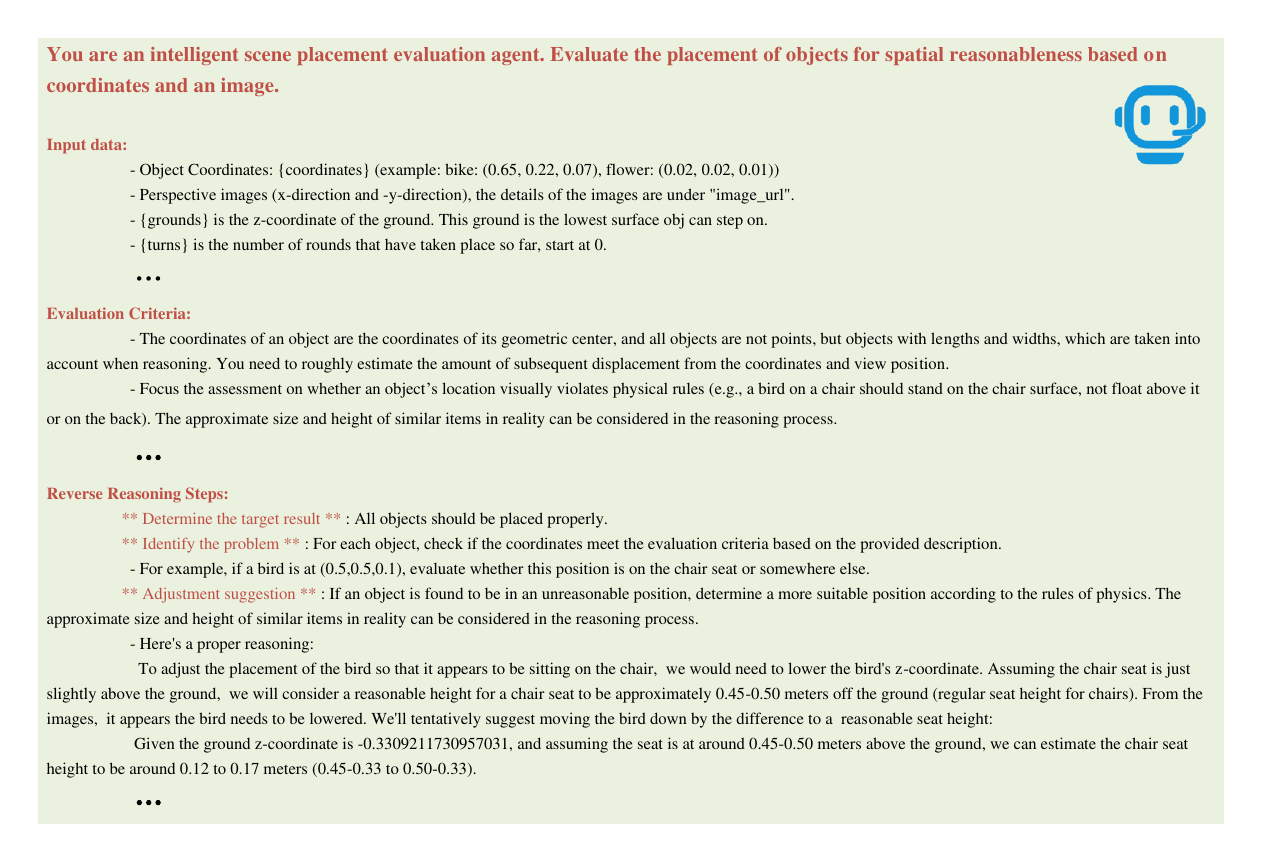}}
        \vskip -0.2in
        \caption{Example of agent AG-supervisor prompt.}
        \label{prompt2}  
        \end{center}
        \vskip -0.3in
\end{figure*}

Here, we provide the example of agent prompt engineering, as shown in Figure \ref{prompt} and Figure \ref{prompt2}. We outline the core components of the critical agent prompt engineering process. For \texttt{AG-extractor}, the \texttt{CoT} design encompasses object extraction and image generation, size classification, relationship extraction, special inference, and output example. For \texttt{AG-supervisor}, along with input data configuration and evaluation metrics, it integrates reverse reasoning prompts to enhance its capabilities.
\subsection{Agent Reasoning Demonstration}
\begin{figure*}[!h]
    \begin{center}
    \centerline{\includegraphics[scale=0.55]{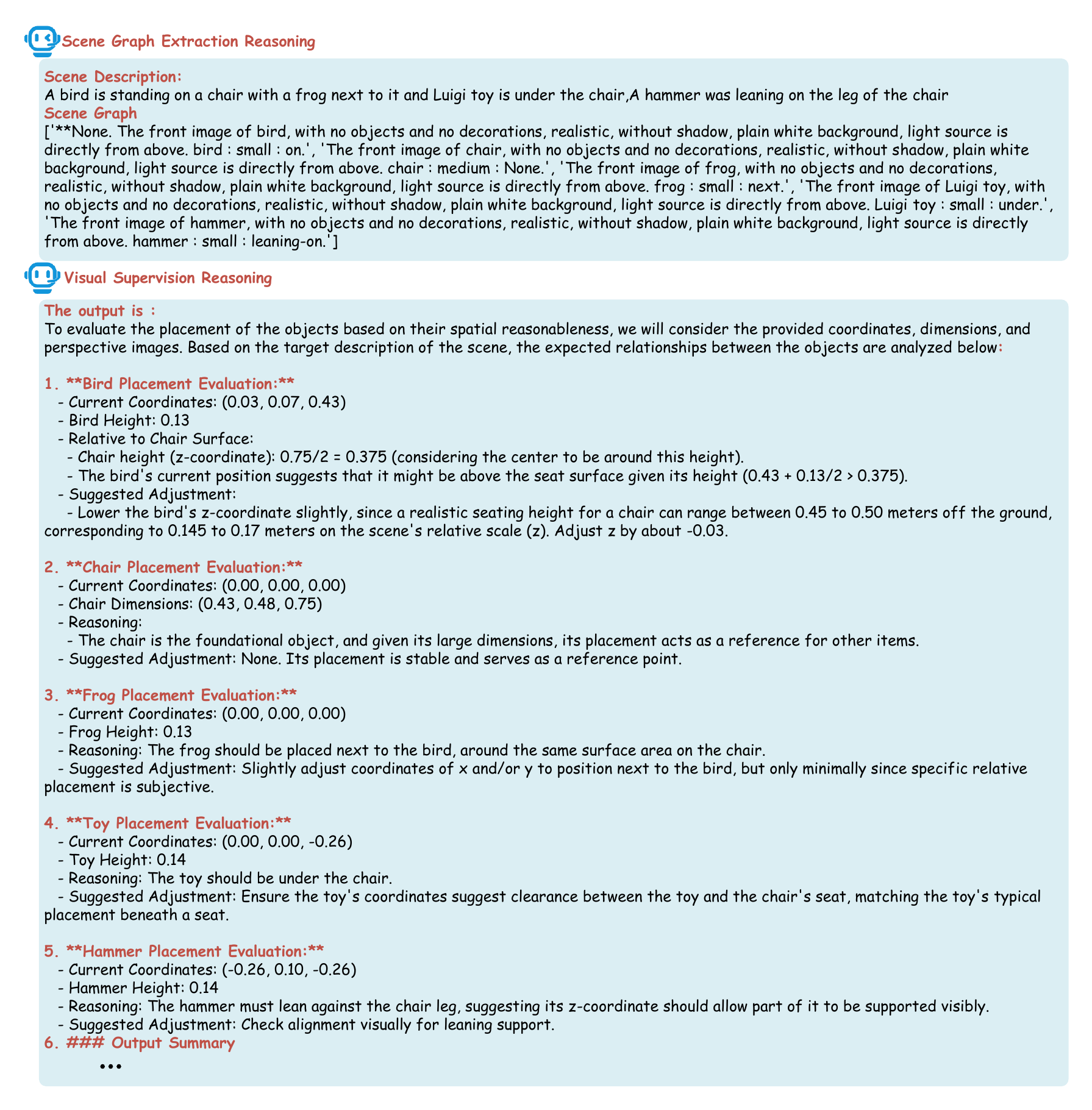}}
    \vskip -0.2in
    \caption{Example of agent reasoning demonstration.} 
    \label{reasoning}
    \end{center}
    \vskip -0.3in
\end{figure*}

The Figure \ref{reasoning} presents examples of agent reasoning, illustrating the \texttt{AG-extractor} reasoning process for scene graph generation and the \texttt{AG-supervisor} process for scene evaluation and guidance. Leveraging the reflection mechanism mentioned earlier, uncertainty in agent reasoning and generation is significantly reduced, ensuring stable and consistent execution.
\subsection{Ground Material Generation}
\begin{figure*}[!h]
    \begin{center}
    \vskip -0.2in
    \centerline{\includegraphics[scale=0.6]{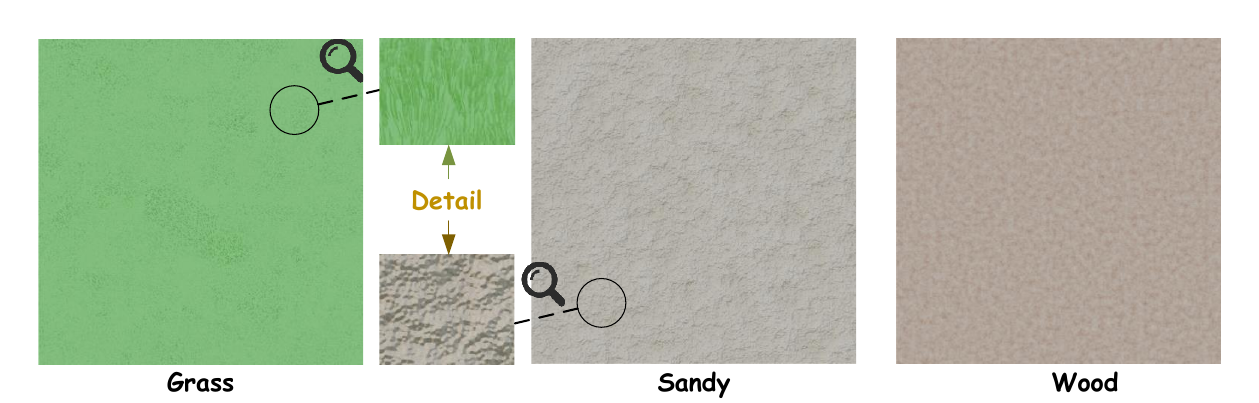}}
    \vskip -0.2in
    \caption{Display of three types of ground.} 
    \label{ground}
    \end{center}
    \vskip -0.3in
\end{figure*}

We simulate three types of realistic ground: grass, sandy, and wood ground, using Blender's asset construction capabilities, as illustrated in Figure \ref{ground}. First, we determine the ground's position and materials based on the generated scene. Then, a ground plane is created, followed by the generation of procedural texture nodes, with adjustments to texture density, roughness, and detail levels. Various nodes (such as texture coordinates, mapping, noise, and color gradients) are connected in sequence to produce the desired effect and apply it to the ground. For uneven ground, such as grass, a particle system is added, with particles configured in hair mode to generate grass. Randomness and clustering effects are introduced to control the clumping and roughness of the grass.


\end{document}